\documentclass[11pt]{article}

\usepackage[margin=1in]{geometry}
\usepackage[T1]{fontenc}
\usepackage[utf8]{inputenc}
\usepackage{lmodern}
\usepackage{microtype}
\usepackage{graphicx}
\usepackage{booktabs}
\usepackage{longtable}
\usepackage{array}
\usepackage{float}
\usepackage{caption}
\usepackage{subcaption}
\usepackage{enumitem}
\usepackage{hyperref}
\usepackage{xcolor}
\usepackage[numbers]{natbib}

\graphicspath{{figures/}}
\setlist[itemize]{topsep=4pt,itemsep=2pt,leftmargin=1.5em}
\setlist[enumerate]{topsep=4pt,itemsep=2pt,leftmargin=1.7em}
\hypersetup{
  colorlinks=true,
  linkcolor=blue!50!black,
  citecolor=blue!50!black,
  urlcolor=blue!60!black,
  pdftitle={Evaluating Large Language Models as Live Strategic Agents},
  pdfauthor={H. C. Ekne}
}

\title{\textbf{Evaluating Large Language Models as Live Strategic Agents:}\\
Provider Performance, Hybrid Decomposition, and Operational Gaps in Timed Risk Play}
\author{H. C. Ekne}
\date{May 10, 2026}

\begin{document}
\maketitle

\begin{abstract}
Static benchmarks capture only part of how large language models behave in practice. Real systems place models inside repeated loops with time limits, formatting constraints, and failure modes. We study this setting in a timed multi-phase Risk environment with explicit victory targets and repeated planning and execution cycles. In a replicated 32-game cross-provider championship under frozen rules, \texttt{gemini-3.1-pro-preview} won 20 of 32 games against \texttt{gpt-5.1}, \texttt{claude-opus-4-7}, and \texttt{kimi-k2.6}, and the pooled winner distribution differs strongly from an equal-strength null ($p \approx 1.5\times10^{-5}$). We then separate planning from execution by standardizing execution on a cheaper Gemini Flash scaffold. Under this design, a pooled 32-game planner bakeoff is consistent with near-equality ($p \approx 0.821$), which indicates that much of the earlier provider spread came from end-to-end system behavior rather than planning alone. To study mechanism, we analyze saved planning and execution traces from the provider championship. Gemini refers to the terminal objective far more often than the other models and increases that focus as victory approaches. Gemini also converts more turns into deep conquest chains, even though it is not the cleanest runtime. These results show that live-agent performance depends on objective tracking, execution conversion, cost, and runtime reliability, and they support evaluating LLMs as components in bounded workflows rather than as isolated benchmark respondents.
\end{abstract}

\section{Introduction}

Most public model comparisons treat LLMs as static respondents. We ask a question, score the answer, and rank the model. That approach is useful, yet it misses how many real systems work. Production systems place models inside bounded workflows with timers, format constraints, repeated phases, and failure modes. A model can look excellent on a benchmark and still underperform once it must act again and again inside a synchronous loop.

This paper studies that gap in a concrete setting: multi-player Risk. We use Risk because it is a compact live-agent benchmark with clear win conditions. The game forces an agent to pursue a long-horizon objective, sequence attacks, respect a constrained action syntax, and adapt to opponents over many turns. Those properties make it a useful stress test for practical LLM behavior.

The central question is straightforward:

\begin{quote}
What changes when we evaluate LLMs as live strategic agents instead of benchmark respondents?
\end{quote}

The paper makes four contributions. It provides a replicated cross-provider live-agent benchmark under a fixed timed protocol. It shows how provider spread changes when planning and execution are separated. It places \texttt{kimi-k2.6} against older closed-model tiers instead of only against the current frontier. It also supplements outcome tables with trace analysis from saved plans and turn summaries.

Three findings drive the paper. Gemini was the strongest tested full-stack stack in the replicated provider championship. In the OpenAI-only lineage tests, \texttt{gpt-5.1} beat newer OpenAI variants under the same live-turn loop. When we standardized execution on a shared cheap Gemini Flash scaffold, the provider spread shrank sharply. That shift suggests that many apparent ``model differences'' are partly system-design differences across planning, execution, timing, and cost.

This paper is best read as a live-agent systems study.

\section{Positioning and Scope}

This paper does not rank general intelligence. It studies a narrower object: \textbf{live-agent behavior under bounded execution constraints}. The main questions are operational. Which model stacks win inside a timed multi-phase loop? How much does that result change once planning and execution are separated? What do those shifts tell us about practical LLM system design?

The study therefore sits between benchmark evaluation and systems evaluation. Risk is the domain. The broader target is any workflow where models must produce constrained actions again and again under a budget.

\section{Related Work}

This paper sits at the intersection of three lines of work. The first line is broad LLM benchmarking. MMLU, BIG-bench, and HELM made public model comparison much more systematic \citep{hendrycks2020mmlu,srivastava2022bigbench,liang2022helm}. Those benchmarks usually evaluate models as respondents. They say less about how models behave inside repeated, timed, stateful loops. That gap matters because many deployed systems fail on formatting, latency, tool use, or recovery from earlier mistakes rather than on one-shot answer quality alone.

The second line is interactive and agentic evaluation. Benchmarks such as AgentBench, GAIA, and $\tau$-bench move closer to realistic usage by measuring multi-step action, tool interaction, and environment feedback \citep{liu2023agentbench,mialon2023gaia,yao2024taubench}. Our work follows that direction. We focus on a synchronous adversarial environment with repeated strategic phases, explicit victory conditions, and hard turn budgets.

The third line studies strategic reasoning in game-like settings. \citet{gandhi2023strategic} examine strategic reasoning with language models in stylized games, and CICERO showed that language plus planning can reach human-level performance in Diplomacy \citep{bakhtin2022diplomacy}. GameBench makes a related case for games as strategic evaluation environments \citep{costarelli2024gamebench}. Our work is closest to this literature. We measure how commercially accessible frontier and near-frontier models behave inside a frozen, costed live-agent loop, and we then ask what changes once we separate planning from execution.

That combination is the gap this paper addresses. Prior work gives us scorecards, agent benchmarks, and game benchmarks. Fewer studies provide a replicated multi-provider competition under fixed budgets while also tracking cost, hybridization, and trace-level mechanisms.

\section{Why Risk Is a Useful Agent Benchmark}

Risk works well as a strategic agent testbed for four simple reasons. It gives each agent a clear terminal objective. In our main runs, that objective was 65\% territory control. It also forces the agent to make several linked decisions inside one turn, such as where to place troops, whether to start an attack chain, when to stop, and how to fortify. The game is adversarial and stateful, so each move depends on board geometry, continent structure, troop counts, and likely responses from opponents. Finally, the repeated timed phases expose operational weakness directly. Timeout drag and fallback behavior become measurable rather than speculative. Together, those features help us separate strategic-sounding language from actual goal-directed play.

\section{Experimental Harness}

\subsection{Environment}

All experiments used the same Risk engine, the same legality assistance from the engine, and the same output grammar. The main live strategic condition used a 90-second planning timer, a 90-second execution turn timer, and a 15-second placement timer. The primary victory condition was 65\% territory control, and seat rotation was enabled. The turn loop itself was broken into pre-turn planning, card trade, troop placement, repeated attacks, and fortify.

\subsection{Core experiment inventory}

\begin{table}[H]
\centering
\caption{Core experiment inventory used in this manuscript.}
\begin{tabular}{>{\raggedright\arraybackslash}p{3.1cm} >{\raggedleft\arraybackslash}p{1.2cm} >{\raggedright\arraybackslash}p{4.2cm} >{\raggedright\arraybackslash}p{4.0cm}}
\toprule
Experiment & Sample & Question & Role in paper \\
\midrule
Cross-provider championship & 32 games & Which full-stack provider wins under the frozen strategic condition? & Main result \\
OpenAI generation ladder & 32 games & Does newer OpenAI release status imply stronger live-agent play? & Recency / prestige result \\
Kimi anchors vs GPT-4.1 / Gemini 2.5 Pro / Sonnet 4 & 16 games each & Where does Kimi sit relative to older closed tiers? & Capability anchoring \\
Gemini Flash cost gate & 15 games & Can a cheaper execution scaffold preserve enough strength? & Cost / system design \\
Flash-exec planner bakeoff & 32 games & Do provider planning differences remain large after execution is standardized? & Decomposition result \\
Provider-32 trace analyses & 946 turns & What planning and execution patterns explain the provider outcome? & Mechanism sections \\
\bottomrule
\end{tabular}
\end{table}

\subsection{Endpoints and statistics}

The primary endpoint in the later experiment program was simple: wins under the configured victory condition. Secondary endpoints included final territory totals, successful and failed attacks, attack-turn rate, fallback counts, invalid move rates, a strategic trace rubric, and estimated API cost where usage logging was available.

For the main omnibus comparisons, we used winner-label permutation tests or equivalent equal-winner Monte Carlo checks depending on the analysis layer. Pairwise win reads used exact binomial logic conditioned on the relevant winner subset. We then supplemented those tests with descriptive execution metrics and with direct analysis of planning and execution traces. This statistical design follows the target the agents were actually given. The agents were told to win the game, so the analysis gives wins the most weight.

\subsection{Reproducibility and artifacts}

Every major run in this paper is backed by saved experiment artifacts, tracked notes, and reusable analysis scripts. Runtime artifacts are stored under the shared \texttt{game\_results} tree, while tracked interpretation lives in the repository documentation. The newer cost and trace analyses are script-backed rather than hand-tabulated.

\section{Main Full-Stack Result: Gemini Wins the Provider Field}

The strongest result in the entire program is the replicated cross-provider championship.

We ran two clean 16-game provider blocks under the same frozen strategic condition and then pooled them. Gemini won 10 of 16 games in the first block and 10 of 16 in the second. The pooled result was Gemini 20/32, OpenAI 6/32, Claude 4/32, and Kimi 2/32.

The pooled winner vector differs strongly from the equal-strength null. The omnibus equal-winner test gives $p \approx 1.5\times10^{-5}$. If we ask how often one specified player would win at least 20 of 32 games in an equal-strength four-player field, the probability is $7.98\times10^{-6}$. Conditioned pairwise winner splits also separate Gemini from each rival. Gemini beat GPT-5.1 by 20--6 with a two-sided $p \approx 0.00936$, Claude by 20--4 with a two-sided $p \approx 0.00154$, and Kimi by 20--2 with a two-sided $p \approx 0.000121$.

\begin{figure}[H]
  \centering
  \includegraphics[width=0.85\linewidth]{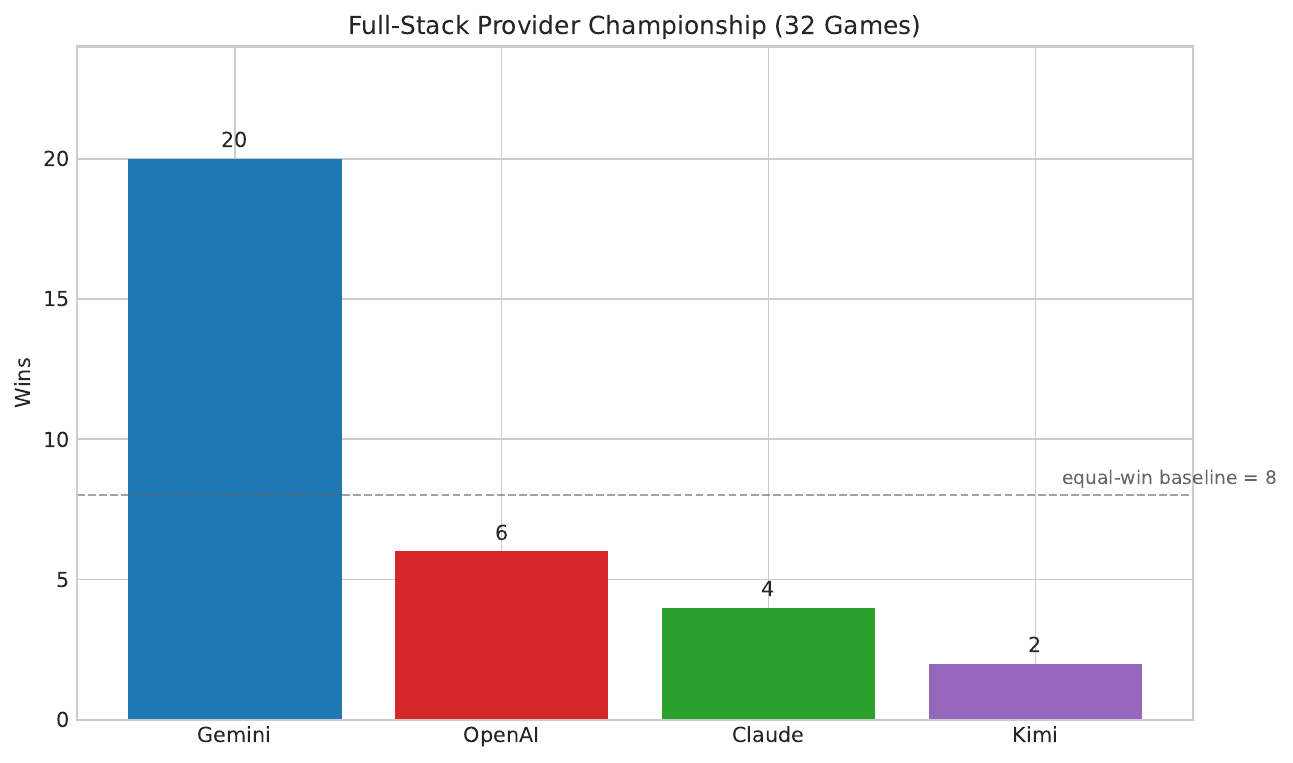}
  \caption{Full-stack provider championship. Pooled wins over 32 games under the frozen full-stack provider setup; Gemini is the only stack with a large replicated lead.}
  \label{fig:providerwins}
\end{figure}

This result is stable across two independent blocks. In this live-agent harness, Gemini 3.1 Pro Preview was the strongest tested full-stack provider representative. We should keep the claim narrow. The result applies to this bounded strategic environment, this rule set, and this timing regime. Even with that scope, the result is strong enough to matter.

\section{Newest or Most Expensive Does Not Mean Best}

The provider result is not the only place where benchmark prestige and deployment reality diverged. In the pooled OpenAI generation ladder, GPT-5.1 emerged as the strongest current OpenAI full-stack baseline in this environment, outperforming newer OpenAI variants in a live-turn condition. That result suggests that synchronous multi-phase loops reward a blend of speed, tactical conversion, and bounded reasoning discipline. Flagship status alone does not guarantee better live play.

The live loop penalizes overlong planning, timeout-driven fallback, brittle formatting, and weak conversion between plan and attack sequence. Once those constraints matter, ``best benchmark model'' and ``best live agent'' become different questions.

\section{Kimi and the Benchmark Mirage}

The Kimi experiments asked a different question. If Kimi is not clearly frontier-class in live play, where does it actually sit? We answered that with three duplicate-team anchor arenas: 2$\times$Kimi vs 2$\times$GPT-4.1, 2$\times$Kimi vs 2$\times$Gemini 2.5 Pro, and 2$\times$Kimi vs 2$\times$Claude Sonnet 4 (2025-05-14).

The pooled picture is coherent. Kimi looks close to GPT-4.1, somewhat below Gemini 2.5 Pro, and competitive with the older Anthropic Sonnet 4 tier. It remains well below the current Gemini 3.1 frontier result.

\begin{figure}[H]
  \centering
  \includegraphics[width=0.88\linewidth]{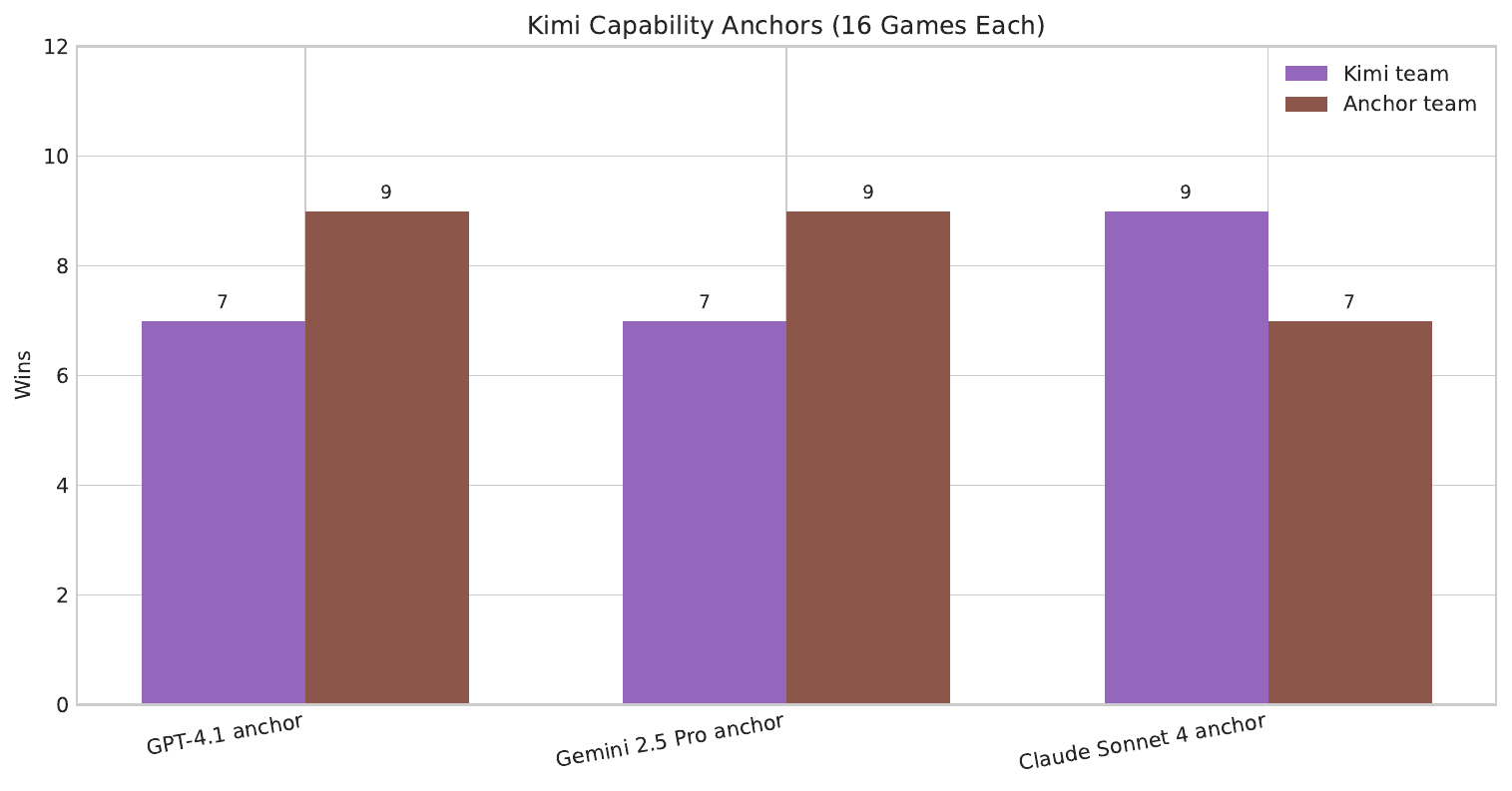}
  \caption{Kimi anchor experiments. Kimi is competitive with older strong closed-model tiers but clearly below the current Gemini 3.1 frontier result.}
  \label{fig:kimianchors}
\end{figure}

Google announced Gemini 2.5 Pro on March 25, 2025, with general availability on June 17, 2025. Moonshot announced Kimi K2.6 on April 21, 2026. So depending on which Gemini release milestone one uses, Kimi 2.6 arrives roughly 10--13 months later.

Yet in this live-agent environment, Kimi only reaches near-parity or modestly trails that older Gemini tier. We should not turn that into a literal calendar-lag theorem. We can still draw a strong practical conclusion. Benchmark-near-parity narratives can overstate real operational parity by a wide margin.

\section{The System Design Turn: Decomposition Beats Monolithic Thinking}

The most important result for builders came after the provider leaderboard question had largely been answered. We then asked a different question:

\begin{quote}
Can we make the benchmark agent materially cheaper without giving away too much strength?
\end{quote}

That question led to a Gemini execution cost gate with three stacks: \texttt{gemini-3.1-pro-full}, \texttt{gemini-3-flash-full}, and \texttt{gemini-3.1-plan\_gemini-3-flash-exec}. In the pooled 15-game cost gate, the hybrid won 8 of 15 games. \texttt{gemini-3.1-pro-full} won 4 and \texttt{gemini-3-flash-full} won 3.

The cost numbers matter just as much. The estimated total cost was about \$6.28 for \texttt{gemini-3.1-pro-full}, \$2.80 for the hybrid, and \$2.08 for \texttt{gemini-3-flash-full}. The hybrid therefore kept most of the strength while cutting cost by more than half.

\begin{figure}[H]
  \centering
  \includegraphics[width=0.88\linewidth]{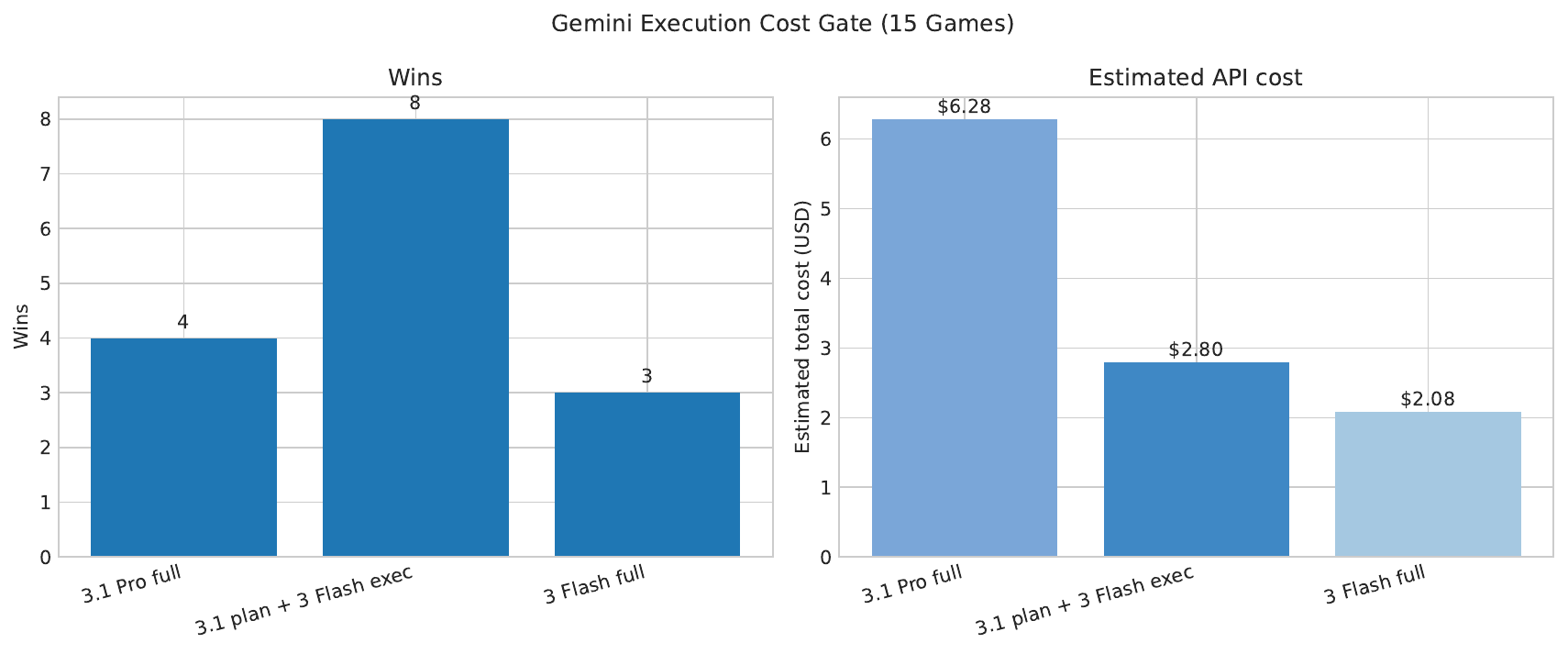}
  \caption{Gemini execution cost gate. The hybrid of Gemini 3.1 planning plus Gemini 3 Flash execution preserves most of the strength while cutting cost materially.}
  \label{fig:costgate}
\end{figure}

This result changes the practical benchmark choice. The best benchmark agent may come from a hybrid design in which a stronger model plans and a cheaper faster model executes.

\section{Planner Rankings Shrink Once Execution Is Standardized}

After locking a shared Flash execution scaffold, we ran a pooled 32-game planner bakeoff with Gemini planning, GPT-5.5 planning, Claude planning, and Kimi planning on the same execution layer. The striking part of this result is the small spread between them.

\begin{figure}[H]
  \centering
  \includegraphics[width=0.88\linewidth]{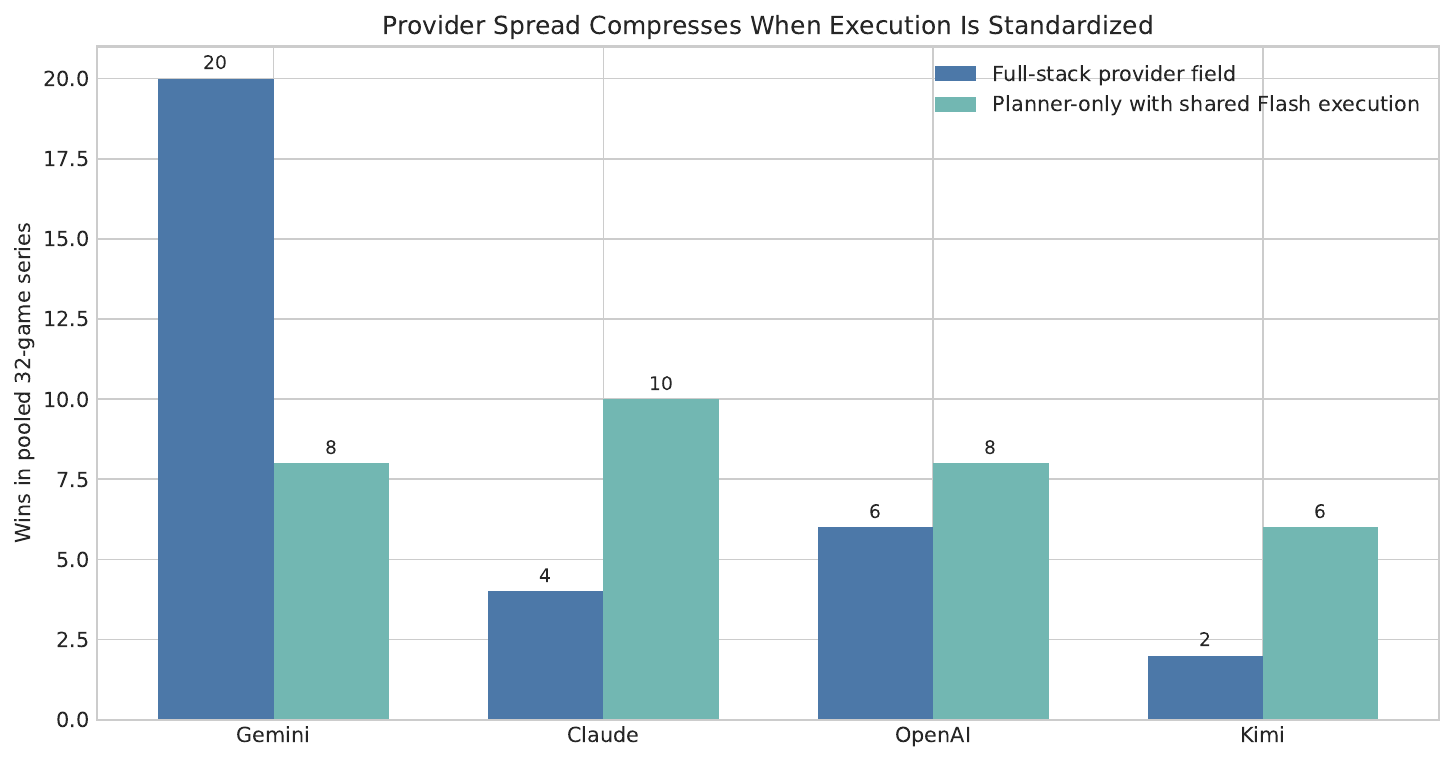}
  \caption{Full-stack spread versus planner-only spread. Once execution is standardized to Gemini Flash, the provider spread compresses sharply.}
  \label{fig:plannercompression}
\end{figure}

Once execution was fixed to the same cheap Gemini Flash layer, the provider spread compressed sharply. The pooled 32-game planner result was consistent with near-equality, with an omnibus $p \approx 0.821$. Even a direct Claude-vs-Kimi duplicate-team duel stayed near parity at 9--7, with a two-sided $p \approx 0.804$.

This result has a strong design implication. Much of the earlier provider spread came from end-to-end system behavior. Planning differences still exist, yet they are much smaller than the full-stack leaderboard first suggested.

\section{Mechanism 1: Gemini Tracks the Objective More Explicitly}

The next step was to inspect the saved planning traces directly. We ran a trace-analysis pass over the pooled provider-32 series and looked at the observable \texttt{plan.text} from 946 saved turn summaries. The question was simple: does Gemini keep the terminal objective more visible in its planning traces?

The answer is yes. Gemini used explicit endgame-goal language in 58.5\% of its saved plans. Claude did so in 3.1\%, Kimi in 1.4\%, and GPT-5.1 in 0.4\%. Gemini also used quantified goal language in 54.5\% of its plans. The corresponding shares were 3.5\% for Claude, 0.8\% for GPT-5.1, and 0.5\% for Kimi.

\begin{figure}[H]
  \centering
  \includegraphics[width=0.88\linewidth]{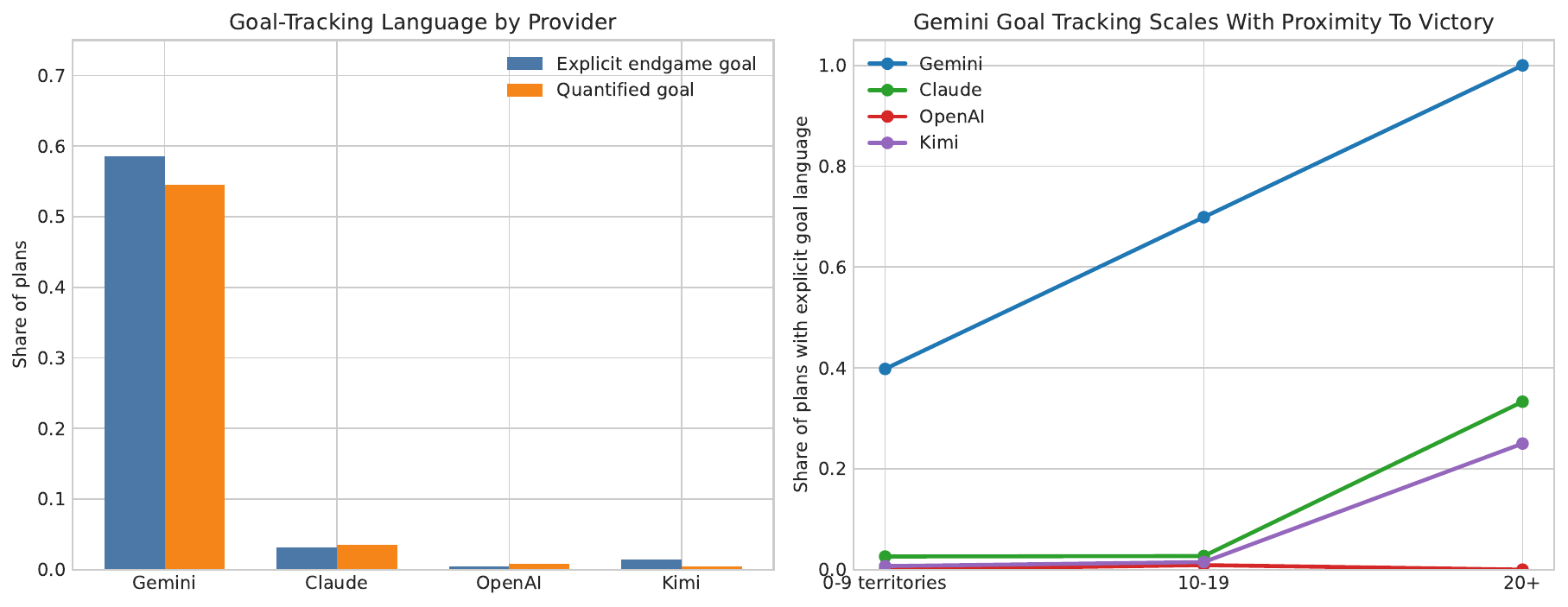}
  \caption{Goal-directedness trace analysis. Gemini references the terminal objective far more often than the other providers and increases that focus as it approaches victory.}
  \label{fig:goaldirectedness}
\end{figure}

This pattern does not come from verbosity alone. Gemini remains the clear outlier even after normalizing by plan length. Its share of plans with explicit goal language rises from 39.8\% when it holds 0--9 territories, to 69.9\% in the 10--19 range, to 100\% in turns where it already controls 20 or more territories. That pattern matches what we would expect from a goal-directed live agent that updates its plan as victory gets closer.

Across all providers pooled, turns with explicit endgame-goal language averaged 5.189 territories gained. Turns without such language averaged 4.080. This comparison does not prove causality, yet it supports the mechanism hypothesis.

\section{Mechanism 2: Gemini Converts Turns Better in Execution}

Planning language alone cannot win games. The model still has to turn a fixed turn budget into board control. The execution-trace analysis over the same pooled provider-32 series shows that Gemini is not the cleanest runtime. Claude is cleaner, GPT can still attack aggressively, and Kimi can sometimes be efficient. Gemini wins this comparison because it converts more live turns into deeper conquest chains than the rest of the field.

\begin{figure}[H]
  \centering
  \includegraphics[width=0.84\linewidth]{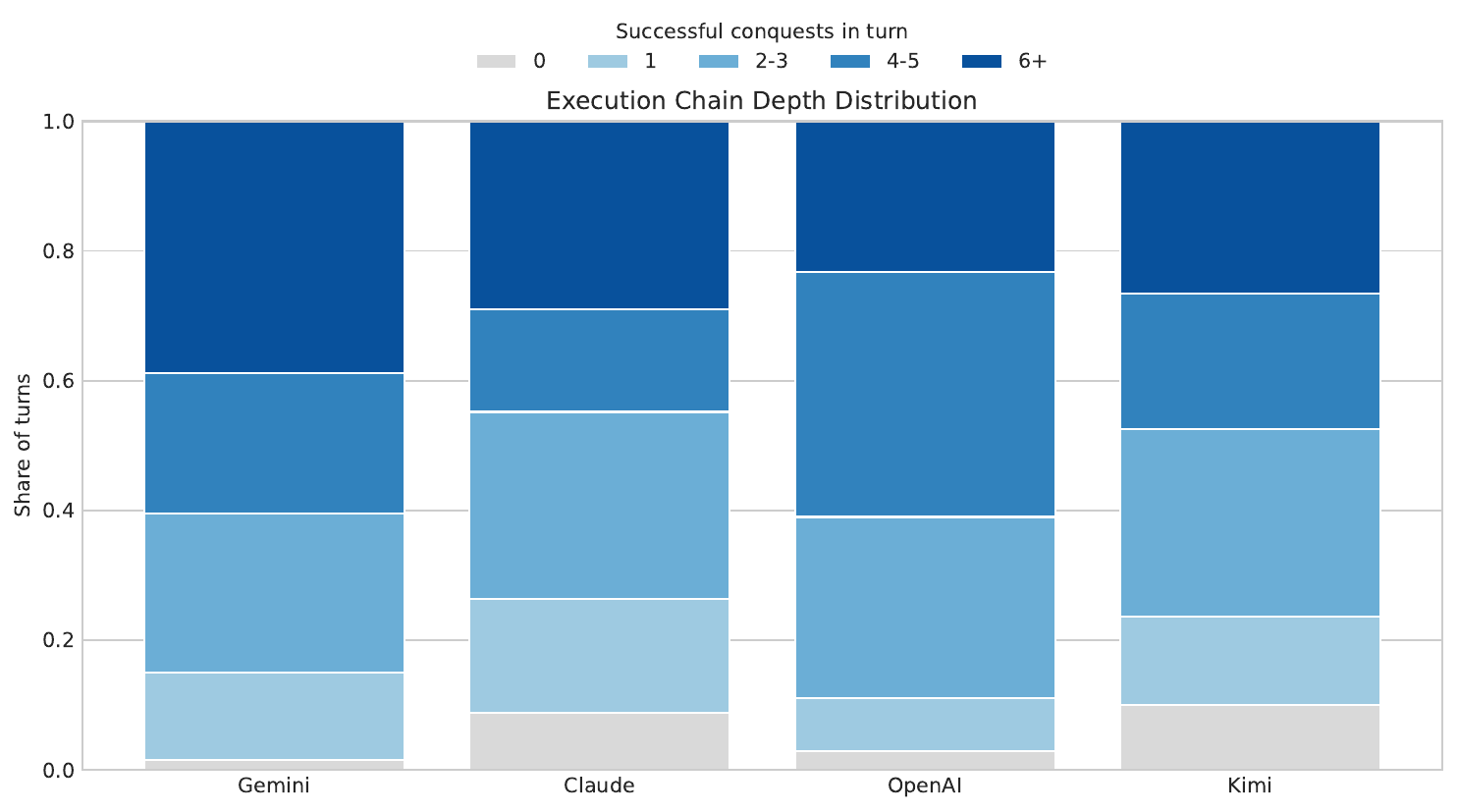}
  \caption{Execution chain depth distribution. Gemini produces deep conquest chains more often than the rest of the provider field.}
  \label{fig:chain}
\end{figure}

Gemini produced 6 or more successful conquests on 38.7\% of its turns. Claude reached that mark on 28.9\% of turns, Kimi on 26.7\%, and GPT-5.1 on 23.4\%. Gemini also had the best midgame territory conversion. When it started a turn with 10--19 territories, it gained 5.363 territories per turn on average. The corresponding values were 4.500 for Kimi, 4.150 for GPT-5.1, and 4.068 for Claude.

\begin{figure}[H]
  \centering
  \includegraphics[width=0.88\linewidth]{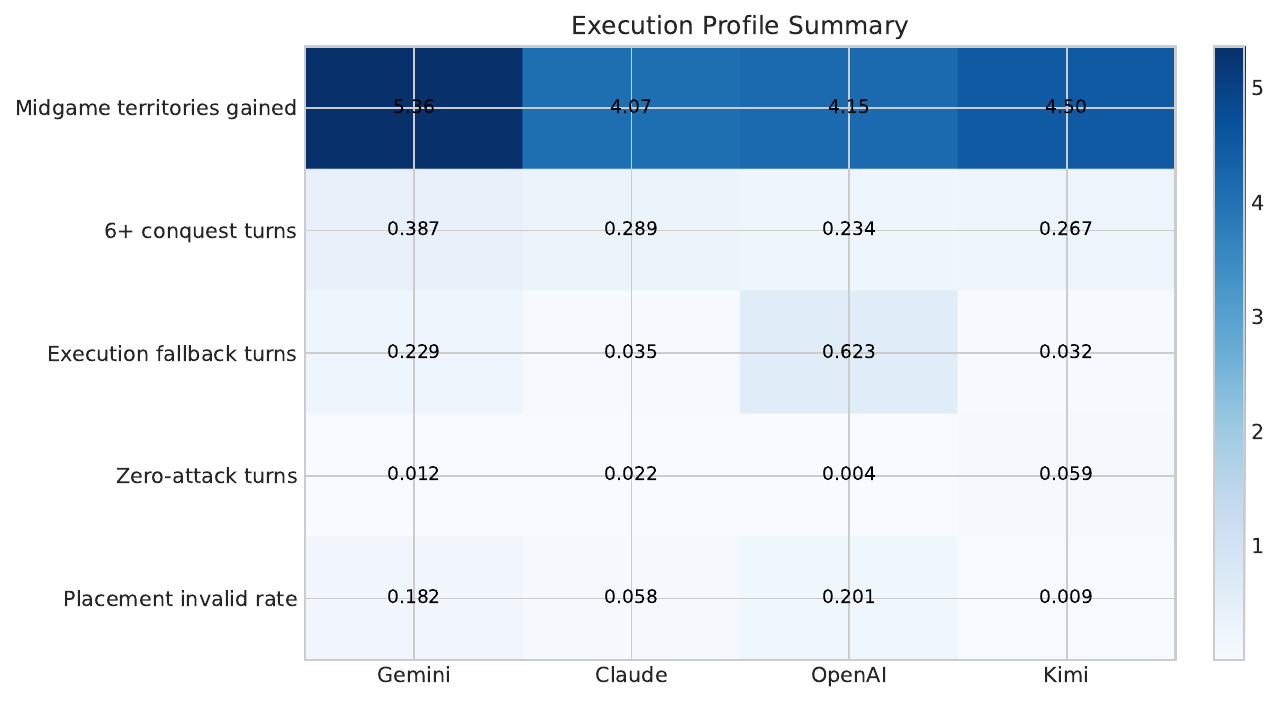}
  \caption{Execution profile summary. Gemini is not the cleanest runtime. It still combines acceptable reliability with the strongest midgame conversion.}
  \label{fig:profile}
\end{figure}

This result clarifies the earlier leaderboard. Gemini did not win because everyone else failed outright. Claude was cleaner. GPT could still push attacks. Kimi often stayed cheap and viable. Gemini won because it paired acceptable reliability with unusually strong conquest conversion.

\section{Implications for Building LLM Systems}

These experiments matter beyond Risk because they point to a different way of evaluating and building LLM systems. A benchmark score does not tell us whether a model will survive a timed loop, maintain output format discipline, keep the objective in view, or convert a plan into repeated high-yield actions. Those factors matter immediately in any multi-step or tool-using system.

The strongest practical scaffold in this project is also instructive. It uses a hybrid design in which a stronger Gemini model handles planning and a cheaper Gemini Flash model handles execution. That pattern is likely to reappear in real systems. Expensive reasoning helps in the stages where abstraction and coordination matter most. Cheaper execution helps in the stages where throughput, latency, and repeated action dominate.

Cost also belongs inside the capability discussion. A model that is accurate but too slow or too expensive to run repeatedly may still be the wrong system component. That is why this project tracked token usage, fallback behavior, and estimated cost whenever usage logging was available. Those measures are part of the engineering problem, not side notes around it.

\section{Limitations}

This study has clear limits. It uses one game domain, one main prompt grammar family, one legality-assistance strategy, and one main timer regime. The planning-trace analysis is observational because it studies visible text rather than hidden reasoning. The ``months behind'' reading for Kimi is anchor-based and approximate. Several planner-only comparisons remain inconclusive once execution is standardized.

Those limits should narrow the interpretation. They do not remove the result. The paper does not claim to solve model ranking in general. It shows that live-agent evaluation can lead to materially different conclusions from the ones suggested by static benchmark culture.

\section{Conclusion}

The main conclusion of this work is simple:

\begin{quote}
Live-agent quality comes from the whole system.
\end{quote}

In this Risk-based strategic benchmark, Gemini was the strongest replicated full-stack provider representative. OpenAI's best live-turn baseline was not its newest flagship-style variant. Kimi sat behind the current frontier despite strong benchmark-adjacent expectations. A cheap planning/execution hybrid produced the strongest practical benchmark scaffold. Once execution was standardized, much of the provider spread in planning collapsed.

That is the core contribution of this study. If we want to understand how LLMs behave in real agent systems, we need to study them as components inside bounded workflows.

\section{Established Claims}

The strongest claims in the paper are narrow and empirical. Gemini is the strongest tested full-stack provider representative in this harness. Benchmark-near-parity claims can overstate live operational parity. Cost-aware planning and execution decomposition can improve deployment value. Standardizing execution compresses much of the apparent provider spread.

Some claims should remain softer. The current evidence does not support universal planner rankings. It does not support exact provider ``months behind'' statements. It does not support broad generalization to every agent domain. It also does not prove that goal-directed language directly causes better play. The manuscript is strongest when it stays close to those boundaries.

\appendix

\section{Appendix: Core Experiment Tables}

\begin{table}[H]
\centering
\caption{Main full-stack provider result used throughout the manuscript.}
\begin{tabular}{lrr}
\toprule
Provider stack & Wins & Win rate \\
\midrule
Gemini 3.1 Pro & 20 & 62.5\% \\
OpenAI representative & 6 & 18.75\% \\
Claude Opus 4.7 & 4 & 12.5\% \\
Kimi K2.6 & 2 & 6.25\% \\
\bottomrule
\end{tabular}
\end{table}

\begin{table}[H]
\centering
\caption{Planner-only bakeoff with execution standardized to Gemini 3 Flash.}
\begin{tabular}{lrr}
\toprule
Planner stack & Wins & Win rate \\
\midrule
Claude Opus 4.7 + Gemini 3 Flash exec & 10 & 31.25\% \\
Gemini 3.1 + Gemini 3 Flash exec & 8 & 25.0\% \\
GPT-5.5 + Gemini 3 Flash exec & 8 & 25.0\% \\
Kimi K2.6 + Gemini 3 Flash exec & 6 & 18.75\% \\
\bottomrule
\end{tabular}
\end{table}

\begin{table}[H]
\centering
\caption{Gemini execution cost gate.}
\begin{tabular}{lrrr}
\toprule
Configuration & Wins & Estimated total cost & Cost per win \\
\midrule
Gemini 3.1 Pro full & 4 & \$6.28 & \$1.57 \\
Gemini 3.1 plan + Gemini 3 Flash exec & 8 & \$2.80 & \$0.35 \\
Gemini 3 Flash full & 3 & \$2.08 & \$0.69 \\
\bottomrule
\end{tabular}
\end{table}

\section{Appendix: Reproducibility and Artifact Links}

\small
\begin{longtable}{>{\raggedright\arraybackslash}p{3.2cm} >{\raggedright\arraybackslash}p{10.5cm}}
\toprule
Artifact category & Location \\
\midrule
\endfirsthead
\toprule
Artifact category & Location \\
\midrule
\endhead
Repository & \url{https://github.com/hcekne/risk-game} \\
Public runtime artifact index & \nolinkurl{docs/article-plans/public_experiment_artifacts.md} \\
Public Dropbox artifact folder & \url{https://www.dropbox.com/scl/fo/s0uxv8uvvd72hbrqnyky8/ACu2RKGnMeufBrLY9US6664?rlkey=yvb70pnmxt1lpd35m531es1w8&dl=0} \\
Latest Dropbox bundle manifest & \url{https://www.dropbox.com/scl/fi/ogu0vyzp9lih533sr4c79/latest_game_results.manifest.txt?rlkey=1rp5izal7kuyhkwrfbm4mvveg&raw=1} \\
Tracked experiment suite index & \nolinkurl{docs/experiment-suites/2026-q2-strategic-tests/README.md} \\
Main provider pooled note & \nolinkurl{docs/experiment-suites/2026-q2-strategic-tests/2026-05-07_cross-provider-frontier-strategic-32-pooled.md} \\
Goal-directedness trace note & \nolinkurl{docs/experiment-suites/2026-q2-strategic-tests/2026-05-10_provider32-goal-directedness-trace-analysis.md} \\
Execution trace note & \nolinkurl{docs/experiment-suites/2026-q2-strategic-tests/2026-05-10_provider32-execution-trace-analysis.md} \\
Gemini Flash cost gate note & \nolinkurl{docs/experiment-suites/2026-q2-strategic-tests/2026-05-09_gemini3-flash-execution-cost-gate.md} \\
Figure generation script & \nolinkurl{scripts/build_live_agent_risk_article_figures.py} \\
PDF manuscript build script & \nolinkurl{scripts/build_live_agent_risk_latex.sh} \\
OCR sanity check script & \nolinkurl{scripts/ocr_check_live_agent_risk_pdf.sh} \\
Shared runtime artifact root & \nolinkurl{/shared-game-results} inside the container workflow; public handoff uses the Dropbox links above \\
\bottomrule
\end{longtable}
\normalsize

\bibliographystyle{plainnat}
\bibliography{references}

\begin{thebibliography}{9}
\providecommand{\natexlab}[1]{#1}
\providecommand{\url}[1]{\texttt{#1}}
\expandafter\ifx\csname urlstyle\endcsname\relax
  \providecommand{\doi}[1]{doi: #1}\else
  \providecommand{\doi}{doi: \begingroup \urlstyle{rm}\Url}\fi

\bibitem[Bakhtin et~al.(2022)Bakhtin, Brown, Dinan, Farina, Flaherty, Fried,
  et~al.]{bakhtin2022diplomacy}
Anton Bakhtin, Noam Brown, Emily Dinan, Gabriele Farina, Colin Flaherty, Daniel
  Fried, et~al.
\newblock Human-level play in the game of diplomacy by combining language
  models with strategic reasoning.
\newblock \emph{Science}, 378\penalty0 (6624):\penalty0 1067--1074, 2022.
\newblock \doi{10.1126/science.ade9097}.

\bibitem[Costarelli et~al.(2024)Costarelli, Allen, Hauksson, Sodunke,
  Hariharan, Cheng, Li, Clymer, and Yadav]{costarelli2024gamebench}
Anthony Costarelli, Mat Allen, Roman Hauksson, Grace Sodunke, Suhas Hariharan,
  Carlson Cheng, Wenjie Li, Joshua Clymer, and Arjun Yadav.
\newblock Gamebench: Evaluating strategic reasoning abilities of llm agents.
\newblock \emph{arXiv preprint arXiv:2406.06613}, 2024.
\newblock URL \url{https://arxiv.org/abs/2406.06613}.

\bibitem[Gandhi et~al.(2023)Gandhi, Sadigh, and Goodman]{gandhi2023strategic}
Kanishk Gandhi, Dorsa Sadigh, and Noah~D. Goodman.
\newblock Strategic reasoning with language models.
\newblock \emph{arXiv preprint arXiv:2305.19165}, 2023.
\newblock URL \url{https://arxiv.org/abs/2305.19165}.

\bibitem[Hendrycks et~al.(2020)Hendrycks, Burns, Basart, Zou, Mazeika, Song,
  and Steinhardt]{hendrycks2020mmlu}
Dan Hendrycks, Collin Burns, Steven Basart, Andy Zou, Mantas Mazeika, Dawn
  Song, and Jacob Steinhardt.
\newblock Measuring massive multitask language understanding.
\newblock \emph{arXiv preprint arXiv:2009.03300}, 2020.
\newblock URL \url{https://arxiv.org/abs/2009.03300}.

\bibitem[Liang et~al.(2022)Liang, Bommasani, Lee, Tsipras, Soylu, Yasunaga,
  et~al.]{liang2022helm}
Percy Liang, Rishi Bommasani, Tony Lee, Dimitris Tsipras, Dilara Soylu,
  Michihiro Yasunaga, et~al.
\newblock Holistic evaluation of language models.
\newblock \emph{arXiv preprint arXiv:2211.09110}, 2022.
\newblock URL \url{https://arxiv.org/abs/2211.09110}.

\bibitem[Liu et~al.(2023)Liu, Yu, Zhang, Xu, Lei, Lai,
  et~al.]{liu2023agentbench}
Xiao Liu, Hao Yu, Hanchen Zhang, Yifan Xu, Xuanyu Lei, Hanyu Lai, et~al.
\newblock Agentbench: Evaluating llms as agents.
\newblock \emph{arXiv preprint arXiv:2308.03688}, 2023.
\newblock URL \url{https://arxiv.org/abs/2308.03688}.

\bibitem[Mialon et~al.(2023)Mialon, Fourrier, Swift, Wolf, LeCun, and
  Scialom]{mialon2023gaia}
Gr{\'e}goire Mialon, Cl{\'e}mentine Fourrier, Craig Swift, Thomas Wolf, Yann
  LeCun, and Thomas Scialom.
\newblock {GAIA}: a benchmark for general ai assistants.
\newblock \emph{arXiv preprint arXiv:2311.12983}, 2023.
\newblock URL \url{https://arxiv.org/abs/2311.12983}.

\bibitem[Srivastava et~al.(2022)Srivastava, Rastogi, Rao, Shoeb, Abid, Fisch,
  et~al.]{srivastava2022bigbench}
Aarohi Srivastava, Abhinav Rastogi, Abhishek Rao, Abu Awal~Md Shoeb, Abubakar
  Abid, Adam Fisch, et~al.
\newblock Beyond the imitation game: Quantifying and extrapolating the
  capabilities of language models.
\newblock \emph{arXiv preprint arXiv:2206.04615}, 2022.
\newblock URL \url{https://arxiv.org/abs/2206.04615}.

\bibitem[Yao et~al.(2024)Yao, Shinn, Razavi, and Narasimhan]{yao2024taubench}
Shunyu Yao, Noah Shinn, Pedram Razavi, and Karthik Narasimhan.
\newblock {$\tau$}-bench: A benchmark for tool-agent-user interaction in
  real-world domains.
\newblock \emph{arXiv preprint arXiv:2406.12045}, 2024.
\newblock URL \url{https://arxiv.org/abs/2406.12045}.

\end{thebibliography}

\end{document}